\documentclass[pdflatex,sn-apa,twocolumn]{sn-jnl}
\usepackage{graphicx}%
\usepackage{multirow}%
\usepackage{amsmath,amssymb,amsfonts}%
\usepackage{amsthm}%
\usepackage{mathrsfs}%
\usepackage[title]{appendix}%
\usepackage{xcolor}%
\usepackage{textcomp}%
\usepackage{manyfoot}%
\usepackage{booktabs}%
\usepackage{tabularx}%
\usepackage{algorithm}%
\usepackage{algorithmicx}%
\usepackage{algpseudocode}%
\usepackage{listings}%
\usepackage{tikz}
\usetikzlibrary{arrows.meta,positioning,calc}
\usepackage{pgfplots}
\usepackage{bm}
\usepackage{xurl}

\usepackage[switch]{lineno} %

\setcitestyle{authoryear,round,aysep={,}}
\renewcommand{\footnoterule}{\kern3pt\hrule height 0.2mm width \dimexpr(\textwidth-\columnsep)/2\relax\kern5.5pt}

\begin{document}

\title[]{Consequential Behaviour and Representational Fairness in the Validation of Synthetic Research}

\author[1,2]{\fnm{Florian} \sur{Kutzner}\textsuperscript{$\dagger$}}
\author[1,2]{\fnm{Celina} \sur{Kacperski}}
\author[1]{\fnm{Laura} \sur{de Molière}}
\author[3,4]{\fnm{Edoardo} \sur{Chidichimo}}
\author[3]{\fnm{Min Jun} \sur{Jung}}
\author[3]{\fnm{Felix P. S.} \sur{Wallis}}
\author[3]{\fnm{James K.} \sur{He}}

\affil[1]{\orgname{decision-context}}
\affil[2]{\orgdiv{Seeburg Castle University}, \city{Seekirchen am Wallersee}, \country{Austria}}
\affil[3]{\orgname{Artificial Societies Ltd.}, \city{London}, \country{United Kingdom}}
\affil[4]{\orgname{University of Oxford}, \city{Oxford}, \country{United Kingdom}}

\abstract{Researchers in industry and academia use synthetic survey respondents powered by large language models as substitutes for human samples. These synthetic populations require validation against real-world data, so researchers often address them using ad hoc comparisons with human surveys. Inspired by the intention–behaviour gap in behavioural science, we argue that these validations test the wrong thing for most applied cases where decision makers commission synthetic research to anticipate consequential behaviour. To address this problem, we propose a validation framework with two requirements. First, every validity claim must state its level of correspondence with human data: does the sample predict what the represented people do, which of four diagnostics (location, dispersion, response process and structure) does the validation address, and does the validation compare against experimental effects? Second, researchers must report validity claims for subgroups, since these groups are often the most affected by consequential decisions and aggregate accuracy hides their misrepresentation. Our validation framework operationalises three justice dimensions (distributional, procedural, and recognition) as measurable quantities and treats within-persona counterfactual experiments as a design that itself requires validation. We then apply the framework to electric vehicle charging tariffs, before closing with a reporting checklist that researchers can use to make convincing validity claims.}


\maketitle
{\renewcommand{\thefootnote}{}%
\footnotetext{\hspace*{-8pt}\textit{Author note.} This framework paper is part of a research programme on the validation of synthetic survey populations conducted jointly by decision-context and Artificial Societies. Funding and competing-interest disclosures appear in the Declarations section.\par\textsuperscript{$\dagger$}Corresponding author, kutzner[at]decision-context.com}

\section*{Introduction}

Several companies now offer access to populations of synthetic survey respondents built on large language models (LLMs). Researchers in the social sciences have started using these respondents as pilot samples, substitutes for hard-to-recruit participants, and a way to test experimental designs before fielding \citep{anthis_llm_2025, bail_can_2024,grossmann_ai_2023}. This technique, often called silicon sampling, addresses several bottlenecks of traditional research. In particular, it can reduce months of fieldwork to a few days, though population-scale simulation still consumes substantial compute and energy \citep{international_energy_agency_energy_2025, sarstedt_using_2024}. Synthetic respondents are also both easy to relate to and easy to access. They can be interviewed whenever the researcher pleases, and they adopt a familiar format for practitioners: the persona. 

The persona has been a working tool in user research for decades \citep{cooper_inmates_1999, pruitt_personas_2003}, and synthetic versions slot neatly into the same workflows (Section \ref{sec:personas-degrees-specificity}). However, unlike statistical models that serve similar predictive purposes in marketing and political campaigns \citep{kosinski_private_2013, matz_psychological_2017, nickerson_political_2014}, these personas' ease of use makes erroneous answers hard to spot and accuracy hard to affirm, while they promise the early voice of the customer whose absence innovation research has long blamed for product failures \citep{cooper_invisible_1999}. Synthetic personas can also be duplicated and put into different experimental conditions in ways that are metaphysically impossible for humans (viz. the fundamental problem of causal inference; \citealp{holland_statistics_1986}). For instance, synthetic personas can run as parallel, uncontaminated counterfactuals of themselves, opening a route to estimating treatment-effect heterogeneity at the level of individual profiles rather than only of subgroups. However, no convention currently exists to evaluate these capabilities. 

In a typical synthetic study, validation usually involves prompting a model with a demographic description and comparing its outputs on attitude or opinion items against a human sample. These validation tests focus on aggregate per-item marginal distributions and thus find strong correspondence between models and humans \citep{argyle_out_2023, bisbee_synthetic_2024, santurkar_whose_2023}. Yet, after establishing this correspondence, subsequent validations do little to assess whether synthetic samples can predict human behaviour and avoid behavioural science's infamous intention-behaviour gap \citep{ajzen_theory_1991, sheeran_intentionbehavior_2016}. 

This oversight is problematic, since decisions like accepting a job, buying a heat pump or agreeing to medical treatment all have costs for the individuals who do them, whereas answering a survey item does not. Likewise, stated-preference methods overstate willingness to pay relative to revealed choices \citep{murphy_meta-analysis_2005}, and self-reports of pro-environmental behaviour correlate only moderately with individuals' actual behaviour \citep{kormos_validity_2014}. Thus, agreement with human survey responses cannot establish whether a synthetic sample accurately predicts the consequential behaviour that practitioners most often want to anticipate, nor can it suggest whether the sample preserves the relationship between intention and behaviour in human samples. 

Uses of synthetic samples fall into three categories that differ in how much they demand of the model: exploratory use (Generating hypotheses, checking question wording), substitutive use (standing in for a human sample in a comparison between conditions or groups) and predictive use (forecasting how a population will respond to a policy or a product before it is introduced). Validation practice has not clearly distinguished these categories, and evidence from the first is often cited in support of the third. Recent methodological guidance organises research with language models around classical validity types and scales the evidence required to the ambition of the claim, from using a model as a tool to treating it as a simulator of a human population \citep{lin_six_2025, lin_simulators_2026, lin_workflow_2026}. We take this as our starting point: the framework below specifies what that evidence must contain when the claim concerns consequential behaviour and particular groups of people.

We use this critique to make three contributions to the nascent field of synthetic research. First, we set out the behavioural criterion and diagnostic levels that researchers must use to make validity claims about human data. Second, inspired by research on algorithmic fairness, we treat subgroup specification as a validity requirement rather than a separate ethical consideration \citep{buolamwini_gender_2018, obermeyer_dissecting_2019}. In doing so, we propose an operationalisation of the distributional, procedural, and recognition dimensions of justice, focusing on subgroup error, documentation, and preserved variance. Finally, we describe within-persona counterfactual experiments, the evaluation design they imply, and what validation against this criterion requires. Our final evaluation framework is purposefully agnostic to synthetic respondent construction, whether via prompting, demographic profiles, or interview data. We conclude that all synthetic research requires documentation of a persona-construction procedure, along with an explicit statement of the research's target validity level and its subgroup claims. 

\section{What to validate}
\subsection{Personas and degrees of specificity} \label{sec:personas-degrees-specificity}

The term `persona' is used differently across fields. In the machine learning literature, the term usually refers to the text that a model is conditioned upon, which can range from a single demographic attribute to the transcript of a two-hour interview \citep{hu_quantifying_2024, park_llm_2026}. At its most basic, a model is prompted without any individual information; here, its outputs rely predominantly on its training data and lean toward the views of younger, better-educated, and more liberal respondents \citep{motoki_more_2024, santurkar_whose_2023}. To counter this, most silicon sampling studies provide models with demographic profile information using in-context learning and, on occasion, supervised fine-tuning \citep{argyle_out_2023, bisbee_synthetic_2024, santurkar_whose_2023}. At the upper end of specificity, models can be fed extensive material about individual persons, including interview transcripts with those persons, demonstrating improved accuracy in representing those individuals \citep{park_llm_2026}. Nonetheless, an open question remains: does higher specificity necessarily improve validity? Our framework enables researchers to test this question. 

Here, we caveat that specificity has a legal limit. A persona constructed to represent a named living person, assembled from material about that person, can raise data protection and profiling issues under the General Data Protection Regulation and the EU AI Act. Accordingly, for this paper, we restrict the definition of a persona to one particular synthetic respondent that represents a hypothetical archetype (as named in \citealp{cooper_inmates_1999}), in the sense of a stylised individual with the characteristics of its profile (that is, the specification text supplied to the model). A persona answers questions; it is not an agent that observes and acts in an environment. It may represent an identified individual only where it is built from that person’s data with their consent, and individual-level claims then additionally require criterion data for that person. All other claims in this paper concern sets of personas, compared with the human population whose characteristics they share and reported per stratum. (We use stratum and subgroup interchangeably for a subgroup of the population specified before validation.)

\subsection{Behaviour as the target of prediction} \label{sec:beh-target-pred}
In our framework, the target of interest is a consequential behaviour observed in a defined population under defined conditions. This behaviour may involve adopting an innovation, choosing a medical procedure, switching tariffs, or completing a financial transaction. We concede that self-reports remain valid targets when the research question concerns experience or opinion itself (e.g., well-being or public support). However, what people say and what they do are different measurable quantities (the intention–behaviour gap), meaning a synthetic sample can match one without matching the other. Beliefs, preferences, and perceived constraints are latent states that shape both what a person says and what they do, yet cost, opportunity, and social desirability affect the two differently, and the gap widens when behaviour is costly \citep{ajzen_attitude-behavior_1977, kormos_validity_2014, murphy_meta-analysis_2005}. Therefore, if models tighten the coupling between self-reports and behaviour, we should not expect them to preserve the looser human coupling between report and action. At best, these models would behave as if people did what they said. Early evidence points this way: simulated respondents couple belief and sharing behaviour almost perfectly where humans show moderate associations \citep{choi_overstating_2026}.

This argument implies that synthetic research validation must predict what people do rather than what people say (Figure \ref{fig:1}), a contrarian position relative to current norms in the field. For example, early studies compared model output to survey items \citep{argyle_out_2023, santurkar_whose_2023}, and critiques of these works then examined response biases, option ordering, and prompt sensitivity in simulated questionnaires \citep{dominguez-olmedo_questioning_2024, tjuatja_llms_2024}. Subsequently, some studies tried to move beyond self-report by examining responses to laboratory tasks and economic games \citep{aher_using_2022, binz_foundation_2025, horton_large_2023}, a route with its own lab-to-field gap since behaviour in laboratory games is only moderately related to behaviour in the field (\citealp{galizzi_external_2019}; see Section \ref{sec:the-criterion}). 

Compounding the issue is a marked decline of directly observed behavioural studies in psychology's flagship journals \citep{baumeister_psychology_2007}, with some counts putting the share of studies measuring actual behaviour at 6 to 18\% \citep{dolinski_is_2018}. Because direct observational experiments are expensive, and cheaper online self-report surveys are attractive, pressure to push behaviour out of the discipline now risks besetting synthetic validation, as researchers continue to pursue verbal correspondence instead of reproducing behavioural outcomes. Therefore, to meet the academics, policymakers, and boards who commission synthetic research and care about behavioural outcomes, the field must course-correct and explicitly state whether validations target verbal or behavioural correspondence. We call this difference between a persona's stated dispositions and the observed conduct of the represented people the \textit{persona intention–behaviour gap}, and the comparison of this quantity with the human coupling for the same population and behaviour \textit{the synthetic intention–behaviour gap} (Figure \ref{fig:1}). 

\begin{figure*}[t]
\centering
\resizebox{\textwidth}{!}{%
\begin{tikzpicture}[
  font=\sffamily,
  >={Stealth[length=2.4mm,width=2mm]},
  box/.style={draw, line width=0.6pt, rounded corners=2pt, align=center, inner sep=6pt,
              minimum width=5.6cm, minimum height=1.15cm, fill=white},
  gaplabel/.style={align=center, fill=white, inner sep=3pt},
  hd/.style={font=\sffamily\bfseries\large},
  sub/.style={font=\sffamily\footnotesize\itshape, text=black!60, align=center},
  corr/.style={align=center, fill=white, inner sep=2pt}
]
\node[hd] (says) at (2.9,8.3) {SAYS};
\node[sub, below=0pt of says] {self-report: attitudes, intentions, survey answers};
\node[hd] (does) at (14.2,8.3) {DOES};
\node[sub, below=0pt of does] {behaviour: verbal, physical or economic, decisional};

\node[box] (hsr) at (2.9,6.7)
  {\textbf{Human self-report}\\[1pt] {\footnotesize what the represented people say}};
\node[box] (psr) at (2.9,-0.3)
  {\textbf{Persona self-report}\\[1pt] {\footnotesize simulated survey answers}};
\node[box, line width=1.3pt, minimum height=1.6cm, fill=black!4] (hb) at (14.2,3.2)
    {\textbf{Human behaviour}\\[1pt] {\footnotesize what the represented people do, in the field}\\[0pt] {\footnotesize\itshape the criterion}};

\draw[dashed, ->] (psr.north) -- (hsr.south);
\node[corr, text width=3.9cm, anchor=east] at (2.7,3.2)
  {\textbf{\small Verbal correspondence}\\[1pt] {\footnotesize\itshape does the persona predict what people say?}\\[2pt] {\footnotesize today's default target (L0--L3 on self-report)}};

\draw[->] (hsr.east) -- node[gaplabel] (hgap)
  {\textbf{\small Human intention--behaviour gap}\\[1pt] {\footnotesize\itshape moderate coupling: roughly half of intenders act \citep{sheeran_intentionbehavior_2016}}}
  ([yshift=8mm]hb.west);

\draw[line width=1.5pt, ->] (psr.east) -- node[gaplabel] (pgap)
    {\textbf{\small Persona intention--behaviour gap}\\[1pt] {\footnotesize\itshape behavioural correspondence: does the persona}\\[-1pt]{\footnotesize\itshape predict what people do? -- the criterion-level check (C)}}
    ([yshift=-10mm]hb.west);

\node[gaplabel, draw=black!35, line width=0.4pt, rounded corners=2pt] (sgap) at (8.55,3.2)
  {\textbf{\small Synthetic intention--behaviour gap}\\[1pt]{\footnotesize\itshape persona gap vs.\ human gap}};
\draw[dashed, ->] (hgap.south) -- (sgap.north);
\draw[dashed, ->] (pgap.north) -- (sgap.south);

\end{tikzpicture}}
\caption{Two kinds of correspondence and three kinds of gaps. Verbal correspondence is how synthetic samples are validated today; behavioural correspondence, operationalised as the persona intention--behaviour gap, is the criterion (C) this paper argues for; the synthetic intention--behaviour gap benchmarks the two.}\label{fig:1}
\end{figure*}
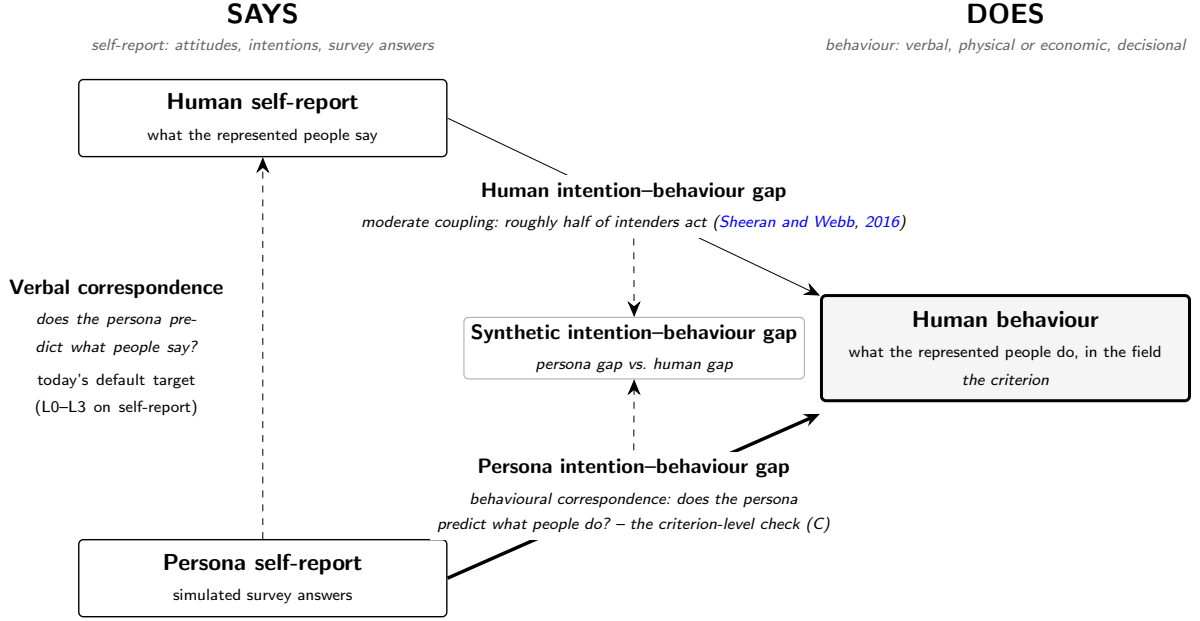

\section{Levels of correspondence between synthetic and human responses}\label{sec:correspondence-between-synthetic-and-human-responses}
Assessing whether we should be convinced by a synthetic study's validation depends on whether the study is exploratory, substitutive (i.e., used in a comparison between conditions or groups), or predictive. This argument follows measurement theory's standard view, where scholars consider validity as a property of a measurement's interpretation and use rather than of the instrument itself \citep{borsboom_concept_2004, cronbach_construct_1955, messick_validity_1995}. Jacobs and Wallach \citeyearpar{jacobs_measurement_2021} apply the same reasoning to fairness in machine learning, describing how algorithms can cause harm when they operationalise a latent construct, such as risk, creditworthiness, or fairness, in ways that diverge from their intended meaning. Since the representativeness of a synthetic sample is a latent construct, what it means to match a human population must therefore differ by application. We implement this logic below through a set of levels, each licensing a different validity claim.

In Table \ref{tab:1}, we define a behavioural criterion (C) that reflects whether the personas in a synthetic study predict the observed behaviour of the group they purport to represent. We separate this criterion from four diagnostic levels (L0 to L3), designed to locate representation failures in a synthetic study. Each level uses a human reference band to describe the expected variation for the same quantity across comparable human samples or repeated measurements. 
The band has two sides because human answering is format- and context-dependent: asked about life satisfaction in general and then about a specific domain, or in the reverse order, people produce correlations of about .32 versus .67 between the same two judgements \citep{schwarz_assimilation_1991}. A persona that answers more consistently than people has therefore failed, and one that holds still where people move will report no effect of a manipulation under test. Accordingly, validation should report uncertainty in the synthetic–human comparison and specify an acceptable discrepancy before inspecting correspondence. Beyond these levels, we use a cross-cutting comparison (E) that focuses on the effects of experimental manipulations and asks whether treatment effects transfer to the synthetic population.

\begin{table*}[t]
\caption{Levels of correspondence between synthetic and human responding}\label{tab:1}
\footnotesize
\setlength{\tabcolsep}{4pt}
\renewcommand{\arraystretch}{1.15}
\begin{tabularx}{\textwidth}{>{\raggedright\arraybackslash}p{16mm} >{\raggedright\arraybackslash}p{15mm} >{\raggedright\arraybackslash}X >{\raggedright\arraybackslash}X >{\raggedright\arraybackslash}X >{\raggedright\arraybackslash}X}
\toprule
Role & Level & Question & Reference point & Illustrative metrics & Pass criterion \\
\midrule
Criterion & \textbf{C} Behavioural & Do personas predict the behaviour of the group of people they purportedly represent? & Observed behaviour of represented persons or strata, in the field if available & Discrimination (AUC, $r$) and calibration per stratum; decision-match rate; synthetic vs.\ human intention–behaviour gap & At or above declared baselines (demographic base rates; unconditioned model); calibration holds within stratum; uncertainty reported \\
\midrule
Diagnostics (to locate failure) & \textbf{L0} Location & Do synthetic marginals match the human population's, per stratum? & Human response marginals for the same target, population and strata & Means, proportions; difference relative to human sampling error & Absolute discrepancy within a prespecified tolerance, with sampling uncertainty reported \\
\cmidrule(lr){2-6}
 & \textbf{L1} Dispersion & Is the spread of synthetic responses human-like, per stratum? & Human response distributions for the same target and strata & Distributional distance (e.g.\ KS); variance ratio; entropy & Equivalence band: neither compressed nor inflated relative to human spread \\
\cmidrule(lr){2-6}
 & \textbf{L2} Response process & Do personas respond to context and format as people do? & Human test–retest bands and context effects & Retest agreement ($\kappa$, ICC); paraphrase stability; sign and size of order, part--whole and format effects & Equivalence band: matches human (in-)\,consistency; reproduces direction and approximate size of human context effects \\
\cmidrule(lr){2-6}
 & \textbf{L3} Structural fidelity & Does inter-item and latent structure resemble the human structure, per stratum? & Human covariance and factor solutions; invariance conventions & Correlation-matrix distance; factor congruence; reliability gap ($\alpha$, $\omega$); measurement invariance across strata & Equivalence band: congruent with human structure and no cleaner; invariance shown for the strata compared \\
\midrule
Cross-cutting & \textbf{E} Experimental & Do manipulations move the synthetic population as they move people? & Human experimental effects for the same population, manipulation and outcome & Effect-sign agreement; ratio of synthetic to human effect size; treatment $\times$ stratum interaction fidelity & Reproduces direction and approximate size of the human effect at whichever level it is read; strongest form reads it at C \\
\bottomrule
\end{tabularx}
\textit{Note.} C is the criterion; L0 to L3 are diagnostics that locate prediction failure. Equivalence bands apply where human data have a characteristic variability that a model can smooth away. Every level is reported per pre-specified stratum (Section \ref{sec:target-group-val}). Row E is not a level: an effect is read at whichever outcome the design targets. Nearest classical analogues: L0 and L3 to the structural aspect and L2 to the substantive aspect of construct validity \citep{messick_validity_1995}, C to criterion validity \citep{cronbach_construct_1955}.
\end{table*}

\subsection{The criterion: Behaviour} \label{sec:the-criterion}
Our behavioural criterion focuses on what people did, recorded in the field where possible. This behavioural evidence comes in three forms: expressive behaviour in the wild (posting, sharing, commenting), physical or economic behaviour (enrolment, purchase, uptake), and high-stakes decisions (incentivised choices, binding recommendations). These measures record actions with consequences for the person or others and, as such, circumvent the low cost of stating an intention in a survey response. To our knowledge, behavioural validation of synthetic samples so far exists only for laboratory tasks and economic games, so two gaps are stacked: one between laboratory and field behaviour \citep{galizzi_external_2019} and one between simulation and human \citep{park_llm_2026}.

To use this evidence, researchers should evaluate behavioural predictions using held-out outcomes unavailable during persona construction, prompt selection, and model tuning. While doing so, their tests should conform to the following rules: (1) baselines must be declared: evaluators should compare personas against demographic base rates and against an unconditioned model, not against chance. (2) calibration must hold within stratum and not only on average, for the reasons outlined in Section \ref{sec:target-group-val}. (3) where self-report is also available, evaluators should compare the personas' intention–behaviour gap to the human intention–behaviour gap observed for the same population and behaviour (Section \ref{sec:beh-target-pred}). Notably, a set of personas for which the two gaps diverge may predict behaviour correctly and still mistranslate interventions that work through intention, such as those captured in message tests or policy pre-tests. Thus, if all three tests hold, the evaluator can claim that the synthetic sample predicts what the stratum of represented people will do, for the given behaviour, population, and model. 

\subsection{Diagnostic levels}
\paragraph{Location (L0)} 
Moving from our behavioural criterion to diagnostics, level 0 assesses whether a synthetic sample lands where a human population does on the quantity of interest. At this level, evaluators should compare the synthetic sample against a population statistic such as a mean or proportion, with the relevant quantity being the difference between synthetic and human estimates relative to the human estimate's sampling error. Previous studies show that models successfully copy, for example, the marginal distributions of American National Election Studies items when conditioned on demographic backstories \citep{argyle_out_2023} and reach aggregate agreement on public-opinion batteries \citep{sun_random_2024}, though performance drops where there are fewer training data for a topic or a population (e.g., non-WEIRD contexts; \citealp{henrich_weirdest_2010}). For an absolute location error, smaller values indicate closer correspondence, although the synthetic mean can depart from the human mean in either direction. Evaluators' strongest claim at this level is that the synthetic sample approximates a population aggregate on this item, for this population, with this model and prompt. They cannot infer anything more about variation, associations, or the effects of interventions.

\paragraph{Dispersion (L1)} 
Level 1 compares the synthetic sample against the spread of the human response distribution (i.e., its variance and a distributional distance), since a sample can have the right average but still misrepresent its population. So far, this is where models fail most consistently. Namely, they compress variance and produce too few responses in the tails, such that a synthetic distribution matches the human mean while remaining too narrow \citep{bisbee_synthetic_2024}. This has been called flattening: a synthetic sample reduces a group's diversity to a single representative position \citep{wang_large_2025}. Evaluators' strongest claims here hold for response spread and minority-position size, subject to the subgroup conditions of Section \ref{sec:target-group-val}.

\paragraph{Response process (L2)} 
Since human survey responses shift with question wording, item order, and response format, at level 2, we test whether the synthetic sample reacts to context and format as people do, using human test–retest bands and documented context effects as reference. \citet{lin_workflow_2026} separates three sources of instability specific to models: variation across repeated runs with an identical prompt, sensitivity to semantically equivalent wordings, and drift across decoding parameters and model versions. L2 takes the same quantities but reads them against the human band, so that a persona that answers more stably than people fails from the other side. The evidence so far is discouraging: LLMs fail to reproduce human response biases \citep{tjuatja_llms_2024}, and their responses are mainly dominated by answer-position and label effects \citep{dominguez-olmedo_questioning_2024}. Therefore, evaluators should assess whether personas' answering processes resemble humans closely enough for context effects to be detectable, a particularly useful trait for applications like message pre-testing.

\paragraph{Structure (L3)} 
Most substantive uses of survey data concern how variables relate to one another, a characteristic that researchers cannot establish by matching question-marginals alone. Thus, level 3 assesses structural fidelity by testing whether the relationships between variables in a synthetic sample correspond to those in human data (i.e. correlations, regression coefficients, factor solutions, and the sign and ordering of predictors). 

To our knowledge, evidence at this level is thin. Barrie and Cerina \citeyearpar{barrie_synthetic_2026} show that synthetic personas distort the covariance structure of belief systems even when univariate distributions look acceptable, so a factor structure or regression estimated on synthetic data does not transfer to humans. Meanwhile, Choi and colleagues \citeyearpar{choi_overstating_2026} find that simulated respondents inflate the correlation between belief and intention to share misinformation, and their regression weights favour attitudinal predictors while discounting the network characteristics that carry most of the association in human data. Associations can also be reproduced for the wrong reasons, for example, from stereotypes in third-party descriptions of a group \citep{bisbee_synthetic_2024, cheng_marked_2023}. Therefore, evaluators should treat a structure cleaner than human data as a failure rather than a success.

\subsection{Experimental correspondence} \label{sec:exp-corr}

Decision makers are often interested in modelling how an intervention will change some population. Accordingly, experimental correspondence (E) measures whether a manipulation affects a synthetic sample as it would with real respondents. Effects can be read at various target outcomes: for example, as a shift in stated support or uptake, comparisons that matter for policy design and for pre-testing experiments. Ashokkumar and colleagues \citeyearpar{ashokkumar_large_2026} validate LLM predictions against a large set of social science experiments, including megastudies, and find strong correlations with human treatment effects but also a systematic overestimation of effect sizes. Likewise, researchers have recovered several classic experimental findings in simulation \citep{aher_using_2022, horton_large_2023}. 

Behavioural research settles for between-person comparison because the same person cannot be run through treatment and control without carry-over or order effects \citep{holland_statistics_1986}; randomisation makes groups comparable in expectation instead \citep{rubin_estimating_1974}, and individual treatment effects are not identified, so heterogeneity must be inferred from comparisons between subjects. A synthetic respondent removes this constraint. Once a persona specification exists, it can be duplicated and each copy assigned to a different condition—a within-persona counterfactual design \citep{lin_illusion_2026}. Holding the persona specification fixed allows repeated draws to estimate a condition contrast for each profile. These contrasts can reveal heterogeneity within the simulation, while their correspondence to human effects requires separate validation.

Four problems complicate the interpretation and validation of these counterfactual contrasts. First, changing a prompt to implement a treatment can change the model’s inference about the respondent, not only the intended manipulation \citep{gui_challenge_2023}; V. Lin and colleagues \citeyearpar{lin_illusion_2026} call this user drift, since the implicit population then differs between conditions and the contrast is confounded. Second, a separate problem is stochastic variation across repeated draws, so two copies of the same persona in the same condition will not answer identically, and the same-condition null distribution is therefore a precondition for interpreting any contrast. \citet{lin_workflow_2026} lists the design controls that such contrasts require: randomised position and format of the manipulated information, explicit constraint of the context a model may otherwise reconstruct around a manipulation, decoding parameters that are fixed or varied as a factor, and analyses that treat repeated draws from one model as clustered rather than independent observations. Third, the individual-level interpretation follows from the profile logic of Section \ref{sec:personas-degrees-specificity}: a within-persona contrast estimates the effect for a profile (i.e., a conditional effect given the specification) which corresponds in human data to the average effect among people matching that profile. Even for consented person-level personas, the contrast remains a model-based estimate of an individual effect, since both potential outcomes cannot be observed for the same person under the same conditions. Fourth, published human estimates are also noisy and subject to publication bias, meaning agreement with meta-analytic estimates or multi-site replications is the stronger validation where such benchmarks exist \citep{camerer_evaluating_2018}. Consequently, the gold standard remains unclear, and evaluators should limit claims to rough magnitudes of treatment effects, for example, to prioritise intervention arms in an experimental design or to screen out interventions. 

Developing a validation procedure is also difficult, as there is no human research equivalent. A starting point could be to compare within-persona counterfactual effects with between-subject effects estimated on human samples matched to the same persona definitions, and to treat systematic divergence as the object of study. Here, two quantities ought to be matched: the average treatment effect in the human sample (viz. an E check) and the ordering of subgroup effects across strata, testing heterogeneity.

\subsection{What each use requires}
The three uses introduced in the Introduction map onto our levels of correspondence; \citet{hullman_this_2026} draw a related line between exploratory and confirmatory use. Exploratory use (hypothesis generation, piloting, checking instruments) requires that L0 to L2 fall within the human bands for each relevant subgroup, and researchers disclose any subgroup for which no comparison is possible. Substitutive use, in which a synthetic sample stands in for a human one in a comparison between conditions or groups, additionally requires L3 invariance across the compared subgroups, demonstrated E correspondence on benchmark manipulations, and a human anchor for calibration or rectification \citep{krsteski_valid_2025}. Predictive use, and any use that allocates resources across groups, requires C for the relevant subgroups at or above declared baselines. Finally, a subgroup that fails invariance or lacks behavioural evidence cannot inherit conclusions drawn from the aggregate.

\section{Target group validity} \label{sec:target-group-val}
\subsection{Aggregate accuracy hides subgroup failure}
Having set out the behavioural criterion and diagnostic levels that researchers should use for their validity claims, we turn to subgroup specification as a validity requirement. In this setting, researchers must first appreciate how an accuracy figure computed over full samples acts as a weighted average dominated by its largest groups. Indeed, a synthetic sample can reproduce a population mean while still being wrong for every subgroup within it, so long as the errors offset. Disaggregated evaluation in machine learning repeatedly documents this pattern, with high error rates in specific intersectional subgroups \citep{buolamwini_gender_2018, obermeyer_dissecting_2019}. As a consequence, disaggregated evaluation has become a reporting standard \citep{mitchell_model_2019}, with calibration requirements within subgroups, and training objectives targeting worst-group performance \citep{sagawa_distributionally_2019}. For language models specifically, Gallegos and colleagues \citeyearpar{gallegos_bias_2024} review available fairness metrics and evaluation datasets and note that most address model output for demographic groups in isolation, which is close to the subgroup logic we propose here.

Three mechanisms produce subgroup failure in synthetic samples. First, models can learn the population-level statistics directly, for example from published polling in their training data, without learning the group-level structure that underpins them. Second, model performance tracks training-data coverage, and coverage tends to be low for very specific subpopulations (e.g., older people, people with low incomes, speakers of less-represented languages, and people with limited digital participation). Unequal adoption of LLM systems, concentrated among younger and better-educated users, points the same way \citep{kacperski_characteristics_2025}. Third, Kumar and colleagues \citeyearpar{kumar_can_2025} show that models trained to answer questions helpfully cannot easily be instructed to mimic incapacity. This matters for consequential behaviour because non-uptake often stems from not knowing an option exists, not understanding it, or not managing the process. Hence, a persona set that cannot represent these states will over-predict uptake in the groups where they are most common. 

Given these failure cases, we argue that researchers must report validity claims about synthetic samples at a subgroup level, following a fixed procedure. Subgroups should be defined before validation and based on the decision context rather than the data. For example, for an energy tariff, relevant partitions could be income band, tenure, dwelling type, and household composition, fixed before inspecting any synthetic data. If sample sizes allow, researchers should also report cells formed by crossing the two or three most decision-relevant attributes, since these intersections are where aggregate-level checks fail \citep{kearns_preventing_2017}. Finally, we encourage researchers to report worst-group performance alongside the average, since this describes the group most likely to be harmed by an error.

Our argument links directly to the validity levels outlined in Section~\ref{sec:correspondence-between-synthetic-and-human-responses}. C reflects calibration in each stratum, as set out in Section \ref{sec:the-criterion}. Meanwhile, at L0 we have a subgroup-specific point estimate, and at L1 we have a subgroup-specific distributional distance, which is where flattening becomes visible, as a model might reproduce the mean of a minority group while removing disagreement within it. At L3 researchers should capture measurement invariance across the strata being compared. Finally, under E, researchers should endeavour to measure subgroup-specific treatment effects (e.g., heterogeneous policy impacts).

While making each claim, researchers must also state the precision of the human benchmark. Subgroup validity is bounded by the size of the human comparison sample, and a subgroup with thirty human respondents cannot support a strong validity claim in either direction. Thus, where human data are scarce, they are better spent on statistically rectifying synthetic estimates than on fine-tuning the model. For instance, Krsteski and colleagues \citeyearpar{krsteski_valid_2025} find that synthesis alone leaves 24 to 86\% bias in population estimates, while rectification with a small human sample brings this below 5\%. Where human data do allow tuning, \citet{lin_simulators_2026} reviews the options, from conditioning on belief networks and richer backstories to fine-tuning on human corpora, and notes that instruction-tuned models already skew towards liberal, higher-income and better-educated respondents relative to their base versions; whichever route is taken, the subgroup report shows for whom it helped. Consequently, although generated responses remain uncorrected, which matters wherever they are used for follow-up questions, a validity guarantee applies to the corrected subgroup estimates. 

\subsection{Justice dimensions as measurable quantities}
To solidify our argument about the importance of subgroup specification in validity claims, we engage with justice frameworks from the literature on machine learning and the energy transition. In the former, frameworks for assessing algorithmic systems' social impacts are numerous, and they converge on principles such as fairness, transparency, and accountability \citep{buckland_future_2025, fjeld_principled_2020, floridi_ai4peopleethical_2018, jobin_global_2019}. However, several reviews criticise them for stopping at the level of principle and rarely specifying measurement or violation rules \citep{hagendorff_ethics_2020, mittelstadt_principles_2019, morley_what_2020}. More recent taxonomies of sociotechnical harm and of generative AI's social impact are more concrete \citep{shelby_sociotechnical_2023, solaiman_evaluating_2024, weidinger_taxonomy_2022}, with impact assessment frameworks that specify a process \citep{metcalf_algorithmic_2021} and require indicators per affected group against an explicit counterfactual \citep{bringmann_quantifying_2026}. It has also been argued that assessing the social impact of a tool whose functioning has not been evaluated is premature, since many harms follow from the lack of functioning \citep{raji_fallacy_2022}; subgroup validity is where synthetic samples meet that functionality question.

However, rather than expand this work, we instead employ a justice framework originally developed for the energy transition whose three dimensions map directly onto the quantities defined in our evaluation framework. These three dimensions include (1) distributional justice concerns – who bears costs and who receives benefits; (2) procedural justice – who participates in decisions and how; and (3) recognition justice – whose situation is acknowledged rather than overlooked \citep{jenkins_energy_2016, sovacool_energy_2015}. Under our definitions, the distributional dimension corresponds to the distribution of prediction error across pre-specified subgroups, since a synthetic sample that is more accurate for some groups than others will encourage decisions better tailored to easily-predicted groups. Likewise, the procedural dimension corresponds to researchers' documentation of the process that produced the synthetic sample; namely, model identity and version, prompt text, sampling temperature, number of draws per persona, the construction of the persona set, and the human benchmark. Without this documentation, external parties cannot reproduce or contest a claim, which becomes a problem when prompt sensitivity is large enough to change conclusions \citep{dominguez-olmedo_questioning_2024, rottger_political_2024}. Datasheets and model cards offer ready templates for this record \citep{gebru_datasheets_2021, mitchell_model_2019}. Finally, the recognition dimension corresponds to preserved within-group variance, as removing dispersion within subgroups (L1) can be classified as representational harm \citep{bender_dangers_2021}. The risks these quantities track have been described from the model side as an identity-essentialisation fallacy, in which demographic labels produce stereotyped and homogenised answers, and as a representation problem that calls for a representation audit before any population is simulated \citep{lin_six_2025, lin_simulators_2026}; the three indicators are our proposal for what such an audit reports. Ultimately, these indicators operationalise selected aspects of justice for validation, though they do not establish that a deployment is necessarily fair.

\section{Application: Charging behaviour in the energy transition}
The framework applies to any consequential decision. We illustrate it with a case from electric vehicle charging, as the distributional issues are explicit and it is a domain in which data about the relevant behavioural outcome (charging patterns) are accessible. Consider an electricity supplier that wants to evaluate a time-of-use tariff intended to shift electric vehicle charging into hours of high renewable generation. The consequential behaviour is the charging start time chosen, observable from meter data. The decision costs the household convenience and involves a financial trade-off. The policy question is whether the tariff shifts charging, by how much, and for which household types. A synthetic evaluation of this design would report, at C, whether the persona set predicts the observed charging start times of each household type at or above its base rate; at L0, whether it reproduces the baseline share of households charging in the target window; at L1, whether it reproduces the spread of start times, which determines how much shifting is available at all; at L3, whether the association between household characteristics and baseline timing matches the observed one; and under E, whether direction and magnitude of the tariff effect match the human estimate, read against meter data and not against stated willingness to shift. 

The subgroup requirement determines what such an evaluation is for. Households differ in whether they can shift charging in the first place. A household with off-street parking and a home charger faces a different decision from one that relies on public infrastructure, and a household in fuel poverty responds to a price signal differently from one for which the saving is negligible \citep{walker_fuel_2012}. If a synthetic sample reproduces the aggregate shift, but overstates it for low-income and renting households, this is a distributional justice issue, as this would support a tariff that delivers its benefits to households with flexibility capital \citep{powells_flexibility_2019}. It is advisable to define subgroups in ways that track behavioural constraints rather than simple demographics; in this case, for example, access to a private charging point is likely the most informative partition. This also limits the number of subgroups.

Within-persona counterfactuals could be used here to examine how household personas respond under tariff and control conditions, providing an estimate of the shift for each household type. The estimate is a hypothesis about heterogeneity that the meter data can then test, which is an appropriate role for a pre-test. This example is meant to showcase what a report structured along the framework may involve, and that missing comparisons at particular levels and for particular subgroups will become visible in such a report.

\section{Reporting checklist}

The framework implies a short list of reporting items. We give them as a checklist because, in our reading of the literature, the main obstacle to comparing validation studies is inconsistency in what is reported rather than disagreement about standards. Studies that use synthetic samples to make claims about human behaviour should report the items given in Table \ref{tab:2}. Items 5 and 6 overlap with the documentation that general reporting guidance for language-model research already asks for \citep{lin_workflow_2026}; the remaining items are specific to validity claims about behaviour and about subgroups. 

\begin{table*}[t]
\caption{Reporting checklist}\label{tab:2}
\footnotesize
\setlength{\tabcolsep}{4pt}
\renewcommand{\arraystretch}{1.15}
\begin{tabularx}{\textwidth}{>{\raggedleft\arraybackslash}p{5mm} >{\raggedright\arraybackslash}p{40mm} >{\raggedright\arraybackslash}X}
\toprule
\# & Item & What must be reported \\
\midrule
1 & Behavioural target and its intended use & The specific behaviour, how it is measured in humans, and whether it is consequential (as discussed in the Introduction); the decision or inference the synthetic sample will inform; the human population it claims to represent; the use claimed (exploratory, substitutive, predictive). If the target is an attitude or intention, this is stated and the claim is not extended to behaviour. \\
2 & Level of correspondence claimed & Which of C, L0--L3, and E is claimed and, for E, the level at which it is read, per Table~\ref{tab:1}, with the corresponding quantity and pass criterion. \\
3 & Human benchmark & Source, size, recruitment and date of the human comparison data; subgroup sizes; effective human sample size per stratum with intervals; any calibration or rectification method and the human anchor used, per stratum. \\
4 & Declared strata & Ex ante list of strata (demographics and intersections, opinion minorities, the target vulnerable or hard-to-reach segment), with rationale; intersections reported where sample sizes permit. \\
5 & Persona construction and provenance & Specificity level; the text supplied to the model; how the persona set was assembled; number of draws per persona; source and legal/ethical basis of any persona-conditioning data; consent design for any tracked behavioural criterion. \\
6 & Model and elicitation specification & Model identity, version and date of access; temperature and other decoding parameters; elicitation formats used and, where L2 is claimed, the formats that were manipulated. \\
7 & Subgroup results against human bands & Results at the claimed level within each pre-specified stratum, each against its human benchmark band; worst-group figure alongside the average; unpopulated cells flagged and retained rather than dropped. \\
8 & Within-subgroup variance ratio & Ratio of synthetic to human variance within each stratum (the L1 measure computed per subgroup). \\
9 & Invariance record & Measurement-invariance method and results across strata and experimental conditions, for any structural (L3) or comparative (E) claim. \\
10 & Criterion evidence or explicit downgrade & Where C is claimed: the behavioural form(s) used (verbal, physical/economic, decisional) and the baselines (demographic base rates, unconditioned model); otherwise an explicit statement that no behavioural criterion was tested and that claims are correspondence with self-report only. \\
11 & Counterfactual null & Where within-persona counterfactuals are used, the same-condition null distribution alongside any treatment contrast. \\
12 & Independence of validation & Who ran the validation; financial and organisational relationships to the evaluated system; pre-registration status and what was locked before results were seen. \\
\bottomrule
\end{tabularx}
\textit{Note.} Items 1, 2, and 7 are the minimum for a study to be assessable; Items 3, 7, 8, 9, and 10 together constitute the validity profile.
\end{table*}

\section{Limitations}
The framework comes with various limitations for which, currently, few or no solutions exist. First, contamination of training data is difficult to separately assess, i.e., a model might reproduce the result of a classic experiment because the paper is in its training data, which yields a correct prediction without any evidence about the underlying process \citep{messeri_artificial_2024}—held-out experiments or using models whose training data pre-dates the event are likely the only mitigations. Second, subgroup analyses are bounded by the availability of human benchmark data at the subgroup level. Much of the attraction of synthetic samples is that they promise access to groups for which human data are scarce, and some of the interest stems from settings where human research data do not exist at all. The concept of algorithmic humility has been proposed as a first rule for such deployments: a system should recognise when it operates at the edges of its training data or outside its competence, and should defer to human review in those cases \citep{buckland_future_2025}. Third, behavioural criteria have their own problems: public posting is performed behaviour, administrative records are shaped by the systems that produce them, and incentivised laboratory tasks are externally mandated and observed. Behaviour is the least bad criterion, but it is certainly not perfect. Fourth, model versions change, and a validity result is related to a specific model at a specific time. Providers update models without notice, and training cutoffs date the attitudes a model can reflect, so that validation evidence has a short shelf life \citep{lin_simulators_2026, lin_workflow_2026}. This argues for reporting model versions and access dates. Finally, and perhaps most crucially, the field still needs a comprehensive set of practical tests to establish which validity requirements synthetic populations meet and where their limitations constrain use. A first battery of tests covering the diagnostic levels, though not the behavioural criterion, is reported in a companion preprint \citep{chidichimo_benchmark_2026}.

One outlook follows from Section \ref{sec:beh-target-pred}. Theories of the attitude–behaviour relation, such as the attitude–behaviour–context model \citep{guagnano_influences_1995} and Campbell's paradigm \citep{kaiser_reviving_2010}, specify when self-report and behaviour converge, namely when the situational cost of acting is neither prohibitive nor negligible. Such theories could serve as a lens for judging when verbal correspondence is sufficient and where behavioural evidence is required, and they suggest that personas may have to be specified as persons in situations rather than as stable characters. A further step in this direction is to place personas as agents inside such a situation and read their behaviour off what they do there, an approach explored so far mostly for self-report-grounded agents \citep{park_llm_2026} and for economic decision tasks \citep{horton_large_2023}, with actual conduct so far checked only against games rather than the field. Where agents interact, validation could also examine whether their patterns of coordination resemble those observed between people \citep{chidichimo_towards_2026}. All the above are questions for later work.

\section{Conclusion}
Synthetic samples generated by language models are being adopted fast. So far, much of the evidence tests whether models answer questionnaires in the same manner as people do, while trying to answer questions concerning consequential behavioural decisions. The gap between what people say and what they do is among the best-documented findings in behavioural science. Claims about consequential behaviour should therefore be validated against observed behaviour and reported separately within groups defined before the analysis.

The framework makes four testable predictions. First, structural fidelity (L3) and behavioural prediction (C) will be at most weakly related across persona sets and models. Second, where verbal fidelity carries behavioural information at all, it will be response-process fidelity (L2) rather than structure (L3) that carries it. Third, aggregate fidelity will overstate worst-subgroup fidelity, and the gap will grow as training-data coverage falls. Fourth, synthetic samples will reproduce the direction of human treatment effects more reliably than their size, and effect correspondence on stated outcomes will not transfer to behavioural outcomes without loss. If structure turned out to predict behaviour beyond process fidelity, the structural programme would be rehabilitated on exactly the criterion we advocate.

We have proposed a framework of predictive validity with a subgroup reporting requirement that follows established practice in machine learning evaluation, an operationalisation of three justice dimensions as quantities that can be read from a results table, and within-persona counterfactual design, which exploits a property of synthetic respondents without counterpart in human research and that itself requires validation. The framework makes claims comparable, and interpretations transparent. The field has spent three years asking how human-like synthetic respondents are on average. The better questions are what they can predict, and for whom.

\section*{Declarations}
\bmhead{Funding} The contribution of the decision-context authors (L.d.M., C.K., and F.K.) was supported by Artificial Societies Ltd. (London, United Kingdom), which funded their working time on this project. Apart from the scientific contributions of the Artificial Societies co-authors in their capacity as researchers, the funder as an organisation had no role in specifying the validation criteria or decision rules and no authority over the decision to publish. The framework is population- and vendor-agnostic, and final authority over content and submission rests with the decision-context authors.

\bmhead{Competing interests} J.K.H., M.J.J., E.C., and F.P.S.W. are founders and/or employees of Artificial Societies Ltd., a company that develops commercial synthetic-audience simulations of the kind this framework is designed to evaluate. L.d.M., C.K., and F.K. received funding from Artificial Societies Ltd. for their working time on this project (see Funding). Indeed, the framework itself requires disclosure of validator independence (Table~\ref{tab:2}, item 12).

\bmhead{Ethics approval} This article proposes a validation framework and reports no new data from human participants; ethics approval was therefore not required.

\bmhead{Data availability} No new data were created or analysed in this article.

\bmhead{Open practices statement} This article proposes a validation framework; no new data or materials were generated and no study was preregistered.

\bmhead{Authors' contributions} Conceptualisation: ALL; Methodology: L.d.M., F.K., C.K.; Writing – original draft: F.K., C.K., L.d.M.; Writing – review and editing: ALL; Funding acquisition: F.K., L.d.M., J.K.H.

\bibliography{decision-context}
\end{document}